\pdfoutput=1
\documentclass[10pt,twocolumn,letterpaper]{article}

\usepackage[pagenumbers]{wacv}

\usepackage{graphicx}
\usepackage{amsmath}
\usepackage{amssymb}
\usepackage{booktabs}

\usepackage[pagebackref,breaklinks,colorlinks]{hyperref}

\usepackage[capitalize]{cleveref}
\crefname{section}{Sec.}{Secs.}
\Crefname{section}{Section}{Sections}
\Crefname{table}{Table}{Tables}
\crefname{table}{Tab.}{Tabs.}

\begin{document}

\title{TimePillars: Temporally-Recurrent
3D LiDAR Object Detection}

\author{Ernesto Lozano Calvo$^1$\thanks{Equal research contribution.} \and
Bernardo Taveira$^1$\textsuperscript{*} \and Fredrik Kahl$^1$\\
\and Niklas Gustafsson$^2$ \and Jonathan Larsson$^2$ \hspace{1cm} Adam Tonderski$^2$ \\
\\
{\tt\small $^1$ Chalmers University of Technology \tiny @chalmers.se \hspace{1cm} \small $^2$ Zenseact \tiny @zenseact.com}}
\maketitle

%%%%%%%%% ABSTRACT
\begin{abstract}
Object detection applied to LiDAR point clouds is a relevant task in robotics, and particularly in autonomous driving. Single frame methods, predominant in the field, exploit information from individual sensor scans. Recent approaches achieve good performance, at relatively low inference time. Nevertheless, given the inherent high sparsity of LiDAR data, these methods struggle in long-range detection (e.g. 200m) which we deem to be critical in achieving safe automation.
Aggregating multiple scans not only leads to a denser point cloud representation, but it also brings time-awareness to the system, and provides information about how the environment is changing. Solutions of this kind, however, are often highly problem-specific, demand careful data processing, and tend not to fulfil runtime requirements.
In this context we propose TimePillars, a temporally-recurrent object detection pipeline which leverages the pillar representation of LiDAR data across time, respecting hardware integration efficiency constraints, and exploiting the diversity and long-range information of the novel Zenseact Open Dataset (ZOD). Through experimentation, we prove the benefits of having recurrency, and show how basic building blocks are enough to achieve robust and efficient results.
\end{abstract}

\newrobustcmd{\B}{\bfseries}
%%%%%%%%% INTRODUCTION

\section{Introduction}
\label{sec:intro}

One of the main challenges in autonomous driving is to create an accurate 3D representation of the surrounding environment, fundamental for reliability and safety. A self-driving vehicle needs to be able to identify objects in its vicinity, such as vehicles and pedestrians, and accurately determine their position, size and rotation. It is common to use deep neural networks operating on LiDAR data to perform this task.

\begin{figure}[t]
  \centering
  \includegraphics[width=0.48\textwidth]{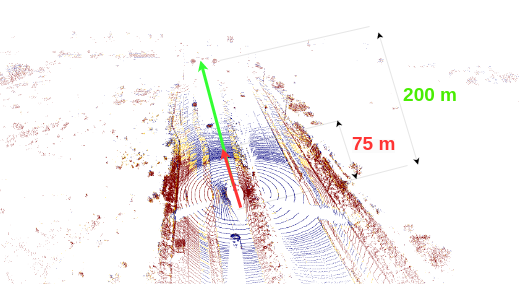}
  \caption{Detection range from LiDAR point cloud data. Depicted in red, relevant established datasets \cite{Kitti,Caesar2019NuScenes:Driving,waymo}, which present annotations up to at most 75 m. In green, long-range information provided by \cite{Alibeigi2023ZenseactDriving}, reaching around 200 m of distance.}
  \label{fig: zod-range}
\end{figure}

The majority of the literature focuses on single-frame methods, i.e.\ on using one sensor scan at a time. This proves to be enough to perform well on classical benchmarks \cite{Kitti,Caesar2019NuScenes:Driving,waymo}, which have objects up to 75~m. However, LiDAR point clouds are inherently sparse, in particular at long ranges. Therefore, we claim that using single scans for long-range detection (e.g.\ up to 200~m, Fig.~\ref{fig: zod-range}) is not sufficient. 

One way to tackle this is with point cloud aggregation, which suggests to concatenate a series of LiDAR scans, yielding a denser input. Nevertheless, aggregating point clouds in this manner is computationally expensive and misses out on the advantages that performing the aggregation within the network can offer. A clear alternative is to use recurrent methods \cite{Chen2022FS-GRU:Sharing,Mccrae3DDATA,Huang2020AnClouds} that accumulate information across time. 

Other techniques to increase the detection range includes using advanced operations such as sparse convolutions, attention modules and 3D convolutions \cite{Fan2023SuperDetection,chen2022mppnet,sun2022swformer, fan2021embracing}. However, compatibility with target hardware is a commonly overlooked aspect of these methods. The hardware used to deploy and to train a neural network are often significantly different. In particular in terms of supported operations and their latency \cite{tensorRT}. Operations such as sparse convolutions or attention are commonly not supported in existing target hardware, e.g.\ the Nvidia Orin DLA. Furthermore, layers like 3D convolutions are often unfeasible due to real-time latency requirements. This enforces the use of simple operations such as 2D convolutions.

In this paper we propose TimePillars, a model that respects the set of supported operations in common target hardware by relying on 2D convolutions, the pillar input representation of \cite{Lang2018PointPillars:Clouds} and a convolutional recurrent unit. 

Ego-motion compensation is applied to the hidden state of the recurrent unit through the use of a single convolution and with the help of auxiliary learning. An ablation study shows the adequacy of using an auxiliary task to ensure the correctness of this operation.

We also study the optimal placement of the recurrent module in the pipeline, and clearly show that locating it between the network's backbone and detection head yields the best performance.

We demonstrate the effectiveness of our approach on the newly released Zenseact Open Dataset (ZOD)~\cite{Alibeigi2023ZenseactDriving}. TimePillars scores noteworthy evaluation performance compared to two PointPillars baselines, one of single-frame nature and another using several aggregated point clouds as input, termed multi-frame (MF) PointPillars. Furthermore, TimePillars achieves significantly higher average precision (AP) on the important cyclist and pedestrian classes, in particular at long range. Lastly, the latency of TimePillars is significantly lower than that of MF PointPillars, making it suitable for a real-time system.  

Our main contributions can be stated as follows:
\begin{itemize}
    \item We propose a novel temporally-recurrent model, TimePillars, which solves the 3D LiDAR object detection task while respecting the set of supported operations on common target hardware.
    \item We show that TimePillars achieves significantly better performance, and in particular at long ranges (up to 200 m), compared to single-frame and multi-frame PointPillars baselines.
    \item To the best of our knowledge, this is the first 3D LiDAR object detection model benchmarked on the novel Zenseact Open Dataset \cite{Alibeigi2023ZenseactDriving}.
\end{itemize}

\paragraph{Limitations. } We restrict our attention to LiDAR data, and no other sensor inputs. Further, we base our method on a single state-of-the-art baseline. Still, we believe our framework
is general in the sense that future improvements to the baseline would translate to overall improved performance of our method.

\section{Related work}

Despite the success of 2D object detection methods \cite{DBLP:yolo,ren2015faster,Liu2015SSD:Detector,tan2020efficientdet}, the need for robust perception has promoted the active exploration of 3D alternatives.
As the first big step in deep learning on raw point clouds, Qi et al. \cite{Qi2016PointNet:Segmentation} proposed a network architecture known as PointNet, which proved to be capable of learning directly on an unordered set of points. Since its release, several approaches followed \cite{Qi2017PointNet++:Space,Shi2018PointRCNN:Cloud}.
Point methods, however, are not efficient enough to face the limitations considered in this work. Projection methods \cite{Yang2019PIXOR:Clouds,Zhou2019End-to-EndClouds} bring the LiDAR data into the (pseudo) image domain directly, where standard image detection architectures can be utilised. Despite being a cheaper alternative, these methods do not extract raw point features, and therefore present a degree of information loss. In this work, we focus on the remaining group, volumetric (voxel-based) methods. These hold a more balanced position in the accuracy-efficiency trade-off.

The current state of the art achieves highly accurate results in the field \cite{Fan2023SuperDetection,erabati2022li3detr,chen2022mppnet,sun2022swformer}. These methods, however, typically benefit from having no constraints on their specifications, which grants them the liberty to employ more sophisticated architectures. Mindful of the hardware limitations that we aim to adhere to, such as restrictions on allowed operations, computational complexity, and memory usage, we filter out baseline-candidates which do not fulfil them. These are models which are either computationally inefficient or possess, 
as key components, operations that are not conducive to integration given the hardware constraints posed, e.g. self-attention layers or sparse convolutions. 
We then analyze the remaining literature, and propose to subdivide relevant work by looking at how sensor data are treated across time. Namely, we introduce single-frame methods, point cloud aggregation techniques, and temporally-aware approaches.

\subsection{Single-frame methods}

VoxelNet \cite{Zhou2017VoxelNet:Detection} was an innovative early proposal which achieved successful results with a voxel-based approach. It was, however, deemed still too slow for real-time applications. SECOND \cite{Yan2018SECOND:Detection} employed sparse-convolutions and achieved a much shorter inference time and improved performance. A substantial change in voxelization philosophy was brought by PointPillars \cite{Lang2018PointPillars:Clouds}, which instead of using regular voxels, point clouds were discretised in vertical columns, defining pillars. 

\subsection{Aggregation of point clouds}

FaF \cite{Luo2018FastNet} demonstrated the benefits in detection accuracy of considering several sensor scans. The implementation used a BEV representation of the LiDAR and analysed the differences between following an Early Fusion scheme (aggregation at the beginning of the network) or a Late Fusion one (successive deeper layers). Later on, IntentNet \cite{Casas2018IntentNet:Data} replaced the frame fusion model, which owned 3D convolutions, by a more efficient module based on 2D convolutions. Most recently, \cite{Fan2023SuperDetection} suggested aggregating only meaningful information. To do that, they introduced the concept of residual points, to name points which change between consecutive frames, i.e.\ which are informative. The combination of these points and previous foreground points (history information) form a richer point cloud. 

\subsection{Temporal awareness}

Differently from aggregation, modelling time dependencies recurrently brings the possibility of preserving relevant knowledge across time, while respecting hardware constraints. After the success of PointPillars, \cite{Mccrae3DDATA} attempted to add a convLSTM in the PointPillars architecture after the backbone and before the detection head. Nevertheless, the results obtained were not improving the overall former architecture if fine-grained pillars were utilized. An alternative, proposed in \cite{Huang2020AnClouds}, chose a 3D sparse data representation. Features were fed into a convLSTM after being processed by a 3D sparse U-Net style backbone. The EGO-motion compensation of the hidden state was performed manually using zero-padding. More recently, FS-GRU \cite{Chen2022FS-GRU:Sharing} extended the application to the motion forecasting task as well, and a convGRU was used to model time dependencies and enable feature sharing. The authors opted for a memory placement before the network's backbone, configuration we prove to be suboptimal via experimentation.

Our approach, TimePillars, takes as starting point the philosophy of PointPillars and extends it in a temporal fusion context. Unlike \cite{Huang2020AnClouds}, we ban the 3D representation of data given its computational cost, and avoid sparse operations considering their inherent incompatibility with hardware accelerators. We deem the ego-motion compensation task to be key, and materialise an efficient way to perform it convolutionally, building upon the intuition behind \cite{Chen2022FS-GRU:Sharing}. However, in contrast to the FS-GRU paper, which just relies on encoded pillar features as hidden state, we prove that utilising further-developed latent information from the backbone adds worthier content to the memory. Our approach, despite requiring a more careful training cycle, demonstrates improved accuracy. Finally, unlike related work seen, we enjoy the novelty of Zenseact Open Dataset~\cite{Alibeigi2023ZenseactDriving} in terms of scale, diversity and long-range LiDAR data.

%%%%%%%%% TIME PILLARS

\section{TimePillars}
\label{sec:timepillars}

In this section, we detail our method, TimePillars, including data processing and neural network architecture.

\subsection{Overview}

\begin{figure*}[ht]
  \centering
  \includegraphics[width=0.8\textwidth]{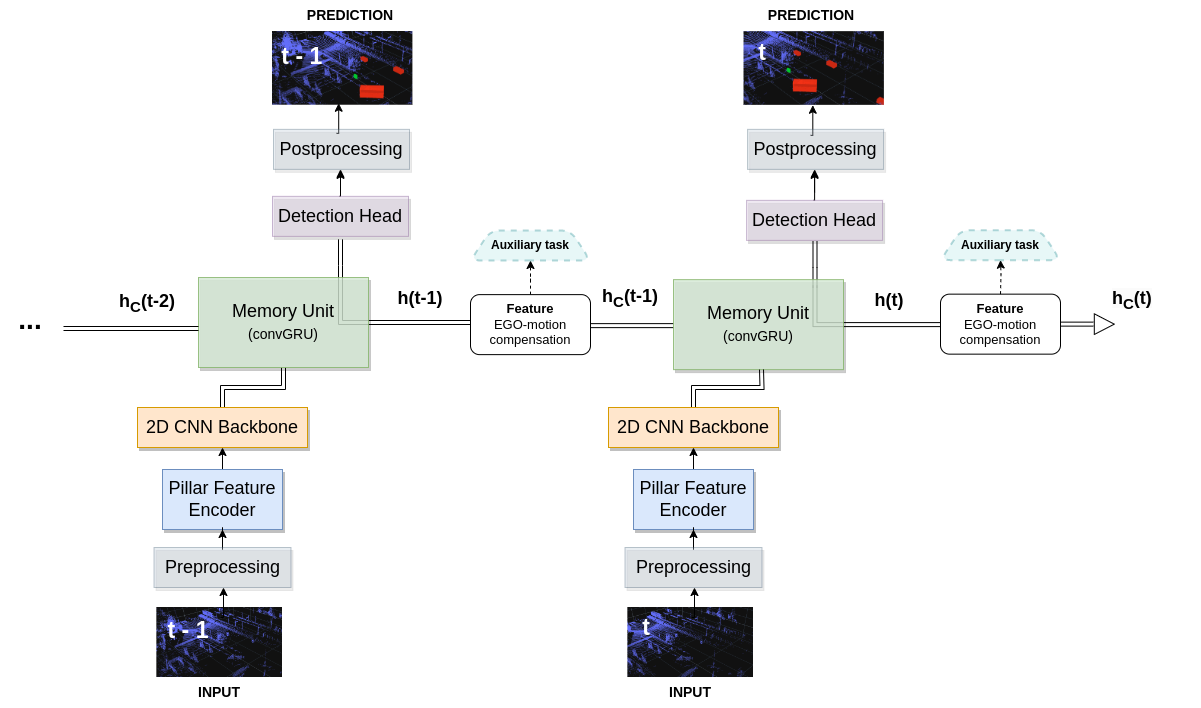}
  \caption{TimePillars pipeline overview. Our method takes LiDAR point clouds as input, and outputs object detection predictions based on past experience (memory) and observations in the present. Preprocessing and postprocessing stages accomodate input raw LiDAR data and detection predictions for feature encoding and bounding boxes extraction, respectively. Valuable information, early encoded as pillars and further developed by a 2D convolutional backbone, is saved by a convGRU-based memory unit. A feature EGO-motion compensation module, guided by auxiliary learning, guarantees memory correctness across time.}
  \label{fig: Model_architecture}
\end{figure*}

We start from the architecture foundations suggested in \cite{Lang2018PointPillars:Clouds} and propose some improvements in the way pillarisation is performed, given its relevance in the process (proximity to raw data). The model is then extended to a multi-frame philosophy. We bet in temporal-awareness as the way to achieve the latter robustly, and establish it by including a convGRU after the backbone, accounting for a long-term system memory. The approach we propose is depicted in Figure~\ref{fig: Model_architecture}, and a detailed description of each of the blocks involved follows.

\subsection{Input preprocessing}

We employ Pillarisation \cite{Lang2018PointPillars:Clouds} to extract a discrete representation of the input point clouds. Unlike voxels, pillars can be seen as vertical column slices, spanning only in the ground dimensions (x and y) while having a fixed height (z) value along the vertical dimension. This approach ensures consistent network input dimensions, eliminates the need for z-axis binning in voxelization while capturing the essential vertical information and enables efficient processing using 2D convolutions. 

A direct issue, however, derives from pillarisation: a large number of pillars from the resulting set are empty, or they own very high sparsity levels (around 97\%, \cite{Lang2018PointPillars:Clouds}). This is a large limitation, commonly faced when dealing with LiDAR point clouds, which the paper authors attempt to tackle by setting a limit in the number of non-empty pillars per sample (denoted as $P$). Such a fix alleviates the problem, but demands the use of pillar sampling or padding to reach the desired limits, operations that can eventually lead to information loss.

In contrast, we apply dynamic voxelization \cite{Zhou2019End-to-EndClouds} with the aim of removing the need to set a predefined number of points per voxel (pillar). Truncation or padding operations are no longer performed for each individual pillar. Instead, data samples are matched to a desired number of total points, $N_t = 200.000$, by treating the points as a whole entity.
 
Denote $D$ as the number of channel dimensions; $N_p$ as the upper limit of points per pillar; and $P$ to be the number of non-empty pillars per sample. Our preprocessing, thanks to Dynamic Voxelization, leads to a tensor of points of shape ($N_t$, $D$), differently from ($D$, $P$, $N_p$) in the original PointPillars paper. This level of abstraction has several beneficial implications. Information loss is minimized (all points available for pillarisation are used) and the resulting embeddings become less stochastic.

\subsection{Model architecture}
We detail the neural network architecture, composed by a pillar encoder, a 2D CNN backbone and a detection head.

\subsubsection{Pillar feature encoder}
The Pillar Feature Encoder maps the (preprocessed) input tensor into a BEV pseudo-image. Given the utilization of Dynamic Voxelization, the simplified PointNet proposed originally is adapted accordingly. The paper applies a 1D convolution to the input, followed by batch normalisation and ReLU, obtaining a $(C, P, N_p)$ tensor, where $C$ stands for the number of channels. Right before the final scatter max layer, max-pooling is applied to the channels, leading to a latent space of shape $(C, P)$. This is where we differ in architecture. Given our tensor was initially encoded as $(N_t, D)$, leading to $(N_t, C)$ after the former layers, we remove the max-pooling operation as it loses adequacy.

Our approach treats all the points as a single entity and therefore there is no longer a need for a dimensionality reduction in the channels. However, the concept of taking the features which maximise the representation remains, and it is implicit in the scatter max layer which follows. Its application leads to a new tensor where feature maximum values are scattered to specific grid positions.

The resulting BEV pseudo-image is of shape $(L, W, C)$, where $L$ and $W$ correspond to the length and width of the grid, respectively. These values are chosen based on the ideal driving perception capabilities, the dataset, and computational constraints. The grid cell size is set to 0.2 m (0.16 m in the original paper) to ease data processing. Aiming for long-range longitudinal detection, we target objects within $[0, 120]$ meters, giving $L = 600$; and consider objects in a width range of $[-40, 40]$ meters, bringing $W = 400$.

\subsubsection{Backbone}

We adhere to the backbone proposed in the original pillars paper, which utilizes a 2D CNN architecture, owing to its favorable depth-efficiency ratio. Three downsampling blocks (Conv2D-BN-ReLU) are utilized to shrink the latent space, and three upsampling blocks with transposed convolutions bring it back upwards, having an output shape of $(\frac{L}{2}, \frac{W}{2}, 6C)$.

\subsubsection{Memory unit}

We model the system's memory as an RNN. In particular we propose the use of a convGRU, which is the convolutional version of a Gated Recurrent Unit. The upgrade from a vanilla RNN prevents from vanishing gradients issues. The adoption of convolution operations respects spatial data and boosts efficiency. Finally, the direct choice of GRU over existing alternatives, e.g.\ LSTM, saves additional computational power and can be seen as a memory regularization technique (hidden state complexity reduction). This is due to the GRU presenting fewer gates, and therefore less trainable parameters.

We reduce the number of required convolutional layers thanks to merging operations of similar nature, consequently making the unit more efficient. In particular, see Eq.(\ref{eq: conv_gates})-(\ref{eq: conv_hidden state}), both the reset $r(t)$ and update $z(t)$ gates share the same input variables (present features, $x(t)$, and saved memory, $h(t-1)$), just owning different learnable parameters ($W_r, b_r, W_z, b_z$). This allows the application of a single convolution of duplicated number of filters via input concatenation. The candidate hidden state, $\tilde{h}(t)$, is obtained, and with it the new memory, $h(t)$.
\begin{equation} \label{eq: conv_gates}
    [r(t)\,, z(t)] = \sigma([W_r\,, W_z)] * [h(t-1), x(t)] + [b_r\,, b_z])
\end{equation}
\begin{equation} \label{eq: conv_candidate_hiddenstate}
    \tilde{h}(t) = \tanh(W_h * [r(t) \cdot h(t-1), x(t)] + b_h)
\end{equation}
\begin{equation} \label{eq: conv_hidden state}
    h(t) = (1 - z(t)) \cdot h(t-1) + z(t) \cdot \tilde{h}(t) .
\end{equation}
The convGRU architecture, through its gating mechanisms, allows the model to adaptively learn which information to forget, which to update, and which new data to incorporate. This enables the capture of long-term dependencies in sequential data.

\subsubsection{Detection head}
We propose a simple modification of SSD \cite{Liu2015SSD:Detector}. Its core philosophy is kept, i.e.\ single-pass and no region proposal. However, we suppress the use of anchor boxes. Instead, a prediction is outputted for each of the cells in the grid directly. The loss of single-cell multi-object detection capabilities, comes with the avoidance of tedious and commonly imprecise tuning of anchor boxes parameters, and inference smoothing.

Linear layers accommodate the individual outputs for classification and localization (location, size and angle) regression. Only the size presents an activation function, and the latter is a ReLU. This is done to prevent it from taking negative values, not being an issue for the other outputs. Alternatively, the classification convolution is followed by a Softmax activation layer which outputs the class probabilities. Furthermore, differently from relevant literature \cite{Lang2018PointPillars:Clouds,Yan2018SECOND:Detection}, we regress the angle output by predicting the heading sine and cosine counterparts independently, and extracting the angle from them. This way, we avoid direct angle regression (damaged by tensor order of magnitude), and the need of a classification head to distinguish front and back of the vehicle.

\subsection{Postprocessing}

To extract the final detection bounding boxes, class score filtering is applied, followed by NMS. The former removes low confidence predictions, while the latter removes redundant boxes associated to a certain object. 2D IoU is used to decide the degree of overlap between two rotated bounding boxes.

\subsection{Feature ego-motion compensation}

The hidden state features outputted by the convGRU are given expressed in the coordinate system of the previous frame. If stored and utilized directly for the computation of the next prediction, a spatial mismatch due to ego-motion would occur.

Different techniques could be applied to perform the transformation. Ideally, already-corrected data would be fed into the network, instead of transforming it on the inside. This is however not our proposal given that it would require, at inference and for every step, resetting the hidden state,  transforming previous clouds and propagating them through the entire network. Not only is this inefficient but also defeats the purpose of using an RNN. The solution in the recurrent context, therefore, demands for the compensation to take place at feature level. This, in turn, makes the hypothetical solution more efficient, but hardens the problem. Classical interpolation methods are used in e.g.~\cite{DBLP:interpolation}. Pseudo-image interpolation can be applied to obtain the features in the transformed coordinate frame. Such a solution, however, may suffer from approximation errors, directly dependant on the polynomial order of the interpolation technique utilized (e.g.\ bilinear, bicubic, etc.).

\subsubsection{Convolutional-based coordinate transform}

In contrast, and inspired by \cite{Chen2022FS-GRU:Sharing}, we suggest to perform the transformation using convolutional operations, and with the help of an auxiliary task. Provided the limited level of detail of the former work, we propose our custom solution to the problem. 

The approach we take provides the network with the needed information to perform the feature transformation via, exclusively, an additional convolutional layer. We start by computing the relative transformation matrix between two consecutive frames, i.e.\ the operation that one would need to perform to successfully transform the features. Next, we extract from it just the 2D information (rotational and translational counterparts) of the matrix:
\begin{equation} 
T_{t-1}^{t} = (T_t)^{-1} \cdot T_{t-1} = \begin{bmatrix} 
    r_{11} & r_{12} & r_{13} & t_{x_{14}} \\
    r_{21} & r_{22} & r_{23} & t_{y_{24}} \\
    r_{31} & r_{32} & r_{33} & t_{z_{34}} \\
    0 & 0 & 0 & 1 \\
\end{bmatrix}
\end{equation}
i.e.\ $r_{11}$, $r_{12}$, $r_{21}$, $r_{22}$ for the rotational part, and $t_{x_{14}}$, $t_{y_{24}}$ for the translational one, respectively.

This simplification avoids the main matrix constants and works in the 2D (pseudo-image) domain, simplifying it from 16 values $(4\times4)$ to just 6. We then flatten the matrix, and extend it to match the shape of the hidden features to compensate $(N_{scans} - 1, L/2, W/2, 6)$. The first dimension denotes the number of frames to transform. This representation makes it suitable for each latent pillar to be concatenated in the channels dimension of the hidden features to be compensated.

\begin{figure}[h]
  \centering
  \includegraphics[width=0.48\textwidth]{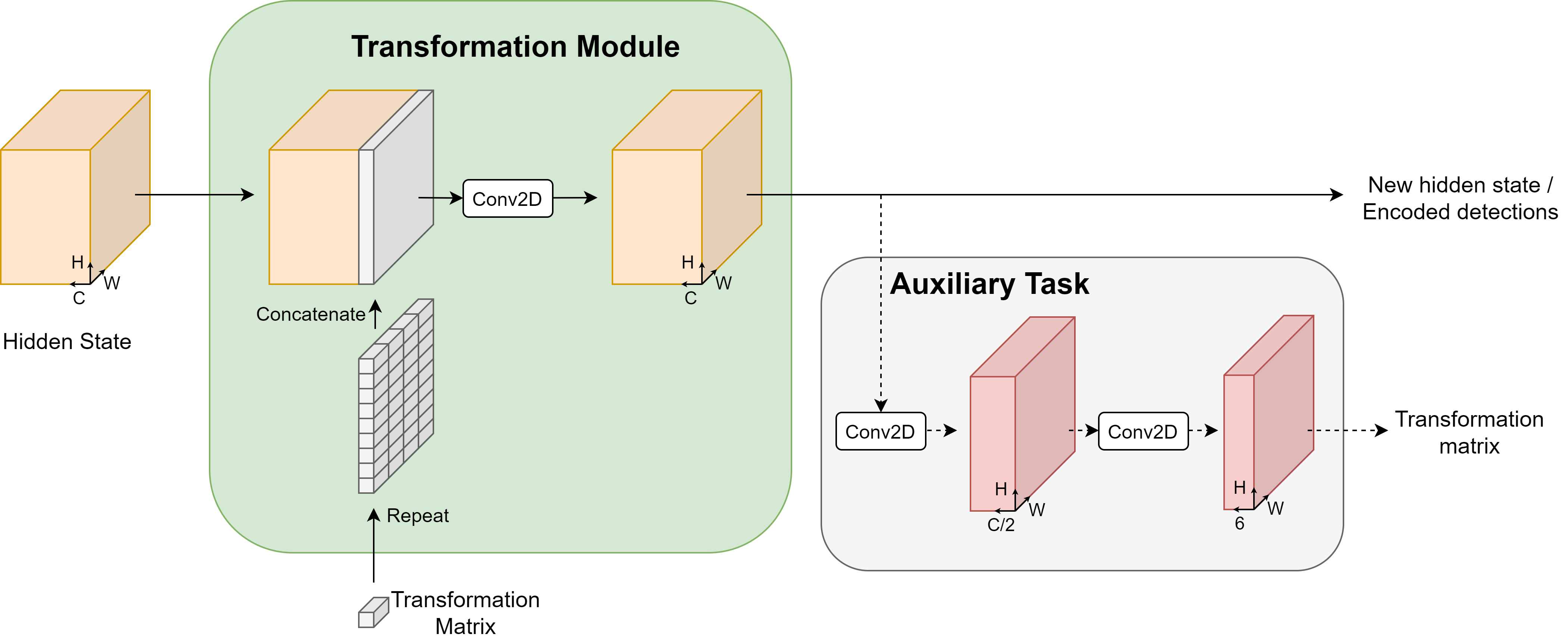}
  \caption{Illustration of both the transformation module and its auxiliary task. A single convolutional layer maps the transformation of the hidden state, provided analytical transformation details in the channel dimension. A lightweight CNN auxiliary task ensures the validity of the transformation module by enforcing transformed features in the output.}
  \label{fig:conv_transformation}
\end{figure}

Finally, the hidden state features are fed into a 2D convolutional layer which fits the transformation process. 
Note a key aspect that derives: the execution of the convolution does not guarantee that the transformation is performed. Channel concatenation just provides the network with extra information about how it could eventually be performed. In this context, we find suitable the use of auxiliary learning. An additional learning objective (coordinate transformation) is added to the training process alongside the primary one (object detection). The design of an auxiliary task, whose aim is to guide the network under supervision through the transformation process, ensures the correctness of the compensation (see Fig.~\ref{fig:conv_transformation}).

The use of the auxiliary task is restricted exclusively to the training process. Once the network has learnt to properly transform the features, it loses adequacy. That is why the task is not considered at inference. Further experimentation, contrasting its impact, can be found in the next section.

%%%%%%%%% EXPERIMENTS

\section{Experiments}
\label{sec:experiments}

In this section we provide extensive experimentation, including the setup, training details, our main results obtained and ablation studies.

\subsection{Setup}

We name the class labels that we focus on, indicate practices which help dealing with class imbalances, and detail the choice of training loss functions.

\subsubsection{Class labels}
The classes we evaluate on are the following: vehicle, vulnerable vehicles (also named cyclist for short) and pedestrian.

Aware of the existing foreground-background class imbalance situation in this field, we perform additional measurements to help with it. Representative class weights scale the learning relevance of each of the classes, based on the frequency of presence in the dataset. Alternatively, we consider a classifier's bias initialisation. Its aim is to speed up convergence, avoiding the network to learn the bias in the early iterations. We do this by ensuring that from start the classification output reflects the imbalance situation, i.e.\ the softmax output corresponds to the frequency of each class.

\subsubsection{Loss functions}
The Focal Loss, first introduced in \cite{Lin2017FocalDetection} is picked as the classification loss. Its choice aims to alleviate the impact of the class imbalance by assigning higher weights to hard, misclassified, examples while down-weighting easy examples. As standard practice, we set the balancing parameter, $\alpha = 0.5$, and the focusing parameter, $\gamma = 2$. 

The Huber loss, or Smooth-L1, \cite{Yan2018SECOND:Detection}, is employed for the regression outputs given its robustness to outliers. We set its quadratic-linear learning threshold value as $\delta = 1.0$ for the location and size, while $\delta = 3.0$ is used for the angle. This is done considering a higher outlier tolerance proved to be of help to encourage the learning of objects' rotation values.

Additionally, we mask out the learning impact of samples which are associated to ground-truth background or unclear labels. Doing this allows us focusing the model's performance on the relevant classes. Moreover, we choose AdamW \cite{Loshchilov2017DecoupledRegularization} as optimiser, to decouple weight decay regularisation from the learning rate setting.

\subsection{TimePillars training cycle}
The dataset's structure affects the recurrent training loop. The frames in the dataset were acquired over a span of two years and across multiple countries, and thus, data linearity is limited. The lack of spatial or temporal relationship among training samples implies that no specific order holds across the entire training loop. 

\begin{figure}[b]
  \centering
  \includegraphics[width=0.475\textwidth]{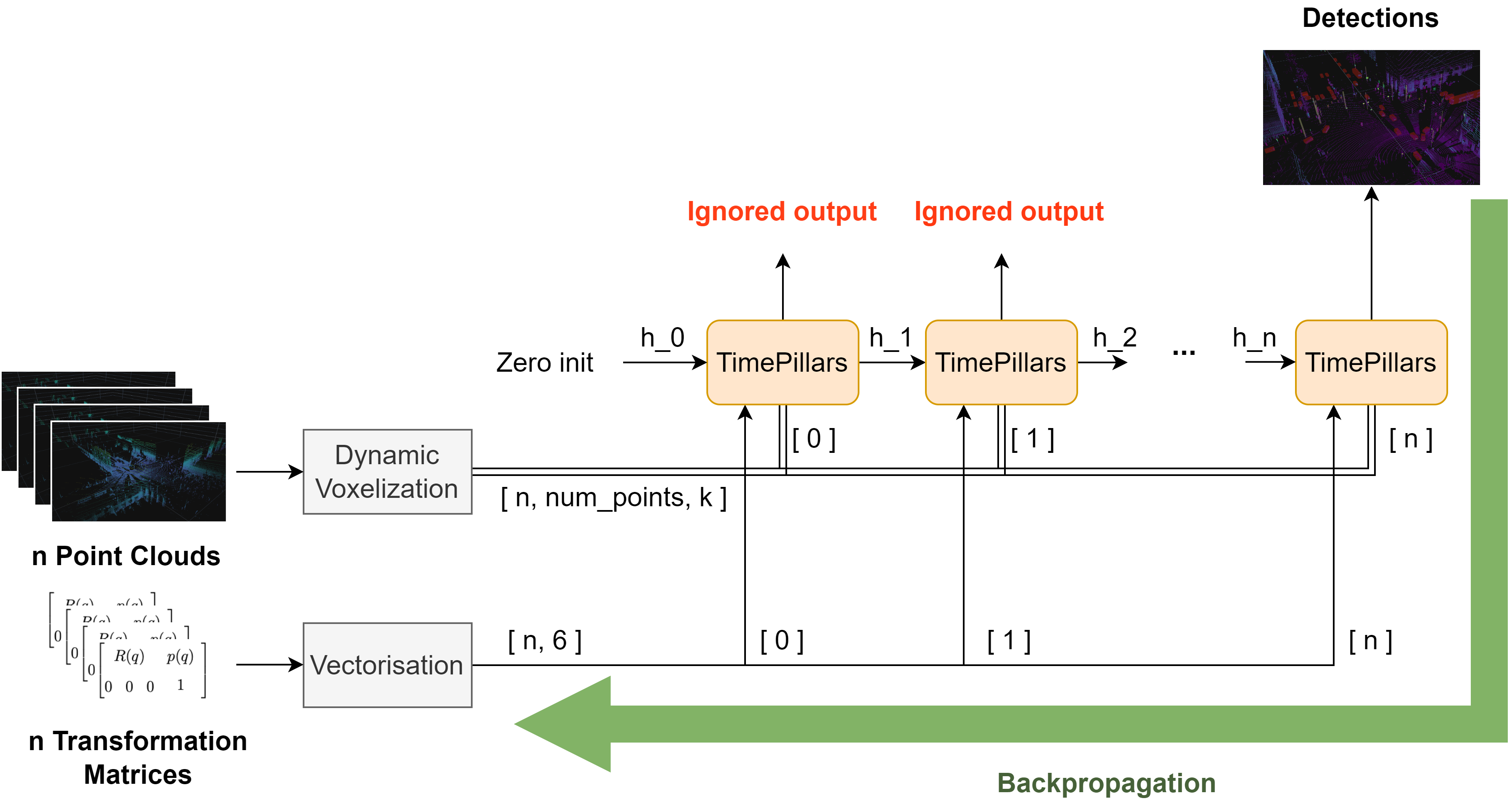}
  \caption{Illustration of a simplified training step. The hidden state is built by running the network for a fixed number of iterations with scans prior to the core frame, ignoring the output. Then, we perform a final pass through the network with the core frame (annotated) as input, and backpropagate to update the weights.}
  \label{fig:train_cycle}
\end{figure}

\begin{figure*}[t]
  \centering
\includegraphics[width=1\textwidth]{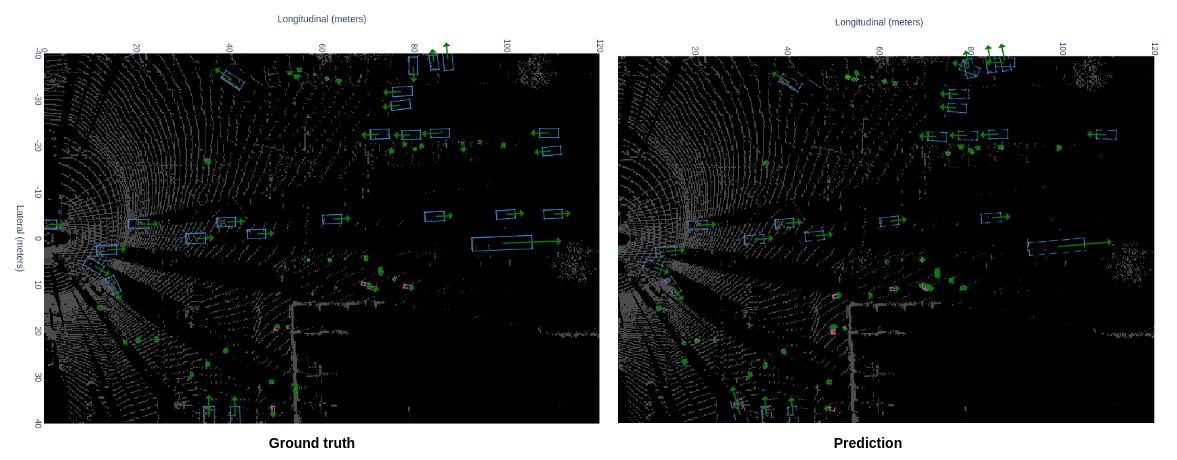}
  \caption{BEV visualization of TimePillars conv-based applied to the driving context. Model predictions are depicted together with the corresponding ground truths. Vehicles in blue, cyclists in red and pedestrians in green. Note the heading direction indicated with an arrow.}
  \label{fig: bevplot}
\end{figure*}
\begin{table*}[t]
\centering
\footnotesize
\begin{tabular}{c c c|c c c c|c c c c|c}
\hline
& & & & & & & & \multicolumn{3}{c|}{Average Precision} & \\
Method & Motion transf. & Scans & NDS & mATE & mASE & mAOE & mAP & Vehicle & Cyclists$^\dagger$ & Pedestrian & f (Hz)\\ \hline
PointPillars* & - & 1 & 0.677 & 0.022 & 0.026 & 0.068 & 0.506 & 0.804 & 0.415 & 0.299 & \B 46.2\\ 
\hline
MF PointPillars* & Preprocessing$^\ddagger$ & 3 & 0.708 & 0.021 & 0.025 & 0.029 & 0.547 & \B 0.809 & 0.454 & 0.379 & 9.1\\ \hline
TimePillars  & Preprocessing$^\ddagger$ & 3 & \B 0.728 & \B 0.018 & \B 0.024 & \B 0.022 & \B 0.577 & 0.802 & \B 0.494 & \B 0.435 & 21.5\\
TimePillars  & Interpolation & 3 & 0.684 & \B 0.018 & \B 0.024 & 0.064 & 0.515 & 0.771 & 0.406 & 0.369 & 20.8\\
TimePillars  & Conv-based & 3 & 0.727 & \B 0.018 & \B 0.024 & 0.029 & \B 0.577 & 0.808 & 0.489 & 0.433 & 21.3\\
\hline
\end{tabular}
\caption{Results of multiple models trained on ZOD frames dataset, using 120 m of range. All models were evaluated with the same number of scans as they were trained with. (*) Custom implementations. ($\dagger$) ``cyclists" is used as a short for vulnerable vehicles, which include motorcycles, wheelchairs, scooter and any human-powered, electric or motorised wheeled object that is used to transport people. ($\ddagger$) By ``preprocessing", we mean that the network does not include any motion transformation layers but instead the data is motion compensated before being fed into the network.}
\label{tab:main_map}
\end{table*}

\begin{table*}[t]
\centering
\footnotesize
\begin{tabular}{c c|c c|c c c|c c c|c c c}
\hline
& & & & \multicolumn{3}{c|}{Vehicle} & \multicolumn{3}{c|}{Vulnerable Vehicle} & \multicolumn{3}{c}{Pedestrian} \\
Method & Scans & NDS & mAP & [0, 50) & [50, 100) & [100, 200+] & [0, 50) & [50, 100) & [100, 200+] & [0, 50) & [50, 100) & [100, 200+] \\ \hline
PointPillars & 1 & 0.657 & 0.475 & 0.878 & 0.753 & 0.498 & 0.469 & 0.393 & 0.036 & 0.404 & 0.211 & 0.000 \\
\hline
TimePillars & 3 & \B 0.723 & \B 0.570 & \B 0.884 & \B 0.776 & \B 0.591 & \B 0.552 & \B 0.521 & \B 0.178 & \B 0.557 & \B 0.350 & \B 0.032 \\
\hline
\end{tabular}
\caption{Distance binning. We evaluate the performance of the proposed model for different detection ranges of interest: from 0 to 50 meters, from 50 to 100 meters, and from 100 to above 200 meters.}
\label{tab: Detect range ablation}
\end{table*}

However, for each core annotated frame, the past 10 scans are available, although not labelled. Given that these past scans cannot be directly used for training due to lack of annotations, we use them to populate the hidden state with pertinent information. This way, we benefit from the massive dataset size, without repeating data or defining a running order (see Figure~\ref{fig:train_cycle}).

Long-term memory stability is fundamental in a recurrent scheme. We observed that this learning context owns the risk of not having long enough sequential annotated data for the network to learn how to preserve relevant knowledge, and how to forget redundant one. We fix this by training on a random number of past scans within a range, this way avoiding scan-overfitting. Randomising epoch steps proves beneficial in the generalisation capabilities of the network.

Transfer learning allowed us to speed up and ease the training process. We first train, from scratch, the building blocks of single-frame nature. Then, the most relevant weights are transferred to train the core of TimePillars. The pillar encoder is frozen for that purpose, as well as the downsampling layers of the backbone, giving freedom for the transferred weights to adapt to the new training context.

\subsection{Results}

Zenseact Open Dataset adapts the metrics presented in nuScenes \cite{Caesar2019NuScenes:Driving} to support long-range data analysis. Table \ref{tab:main_map} holds the main results obtained in this paper, and Figure \ref{fig: bevplot} serves as visualization. TimePillars of different ego-motion transformation nature are compared against our single-frame baseline (PointPillars), and a multi-frame extension of it (aggregation of 3 past scans at the input). All the models contained in the table are trained and evaluated using 120~m of longitudinal range. Metrics show that TimePillars outperforms the other methods presented in most of the categories. The highest overall nuScenes detection score (NDS) is achieved when already-compensated data are fed into the network, as then it does not need to learn the transformation itself. As mentioned previously, this scenario is ideal and posed just for completeness. Our main proposal, convolutional-based, manages to nearly match the performance of the latter, and is around 5\% superior than its direct alternative, based on bilinear interpolation. 
Class-specific AP values show that MF PointPillars presents the highest precision for the predominant class, vehicles, while the conv-based TimePillars surpasses PointPillars related methods in harder classes like cyclists and pedestrians, being the improvement for the second significant. This growth in quality can be attributed to the use of past information, key for classes which are less likely in the dataset and of high sparsity. We also provide, for illustration, inference speeds on the NVIDIA A100. The models are quantized to float16 using TensorRT, but no further optimization techniques are performed. Numbers showcase the existing trade-off between accuracy and runtime. The single-frame approach, intuitively, runs faster than the methods which use several sensor scans, but our recurrent method proves to be twice faster than direct aggregation methods.

\subsection{Ablation studies}

Relevant ablation studies are performed, namely we study the recurrent module's placement, we prove the adequacy of having an auxiliary task, and we explore the longitudinal detection range.

\subsubsection{Memory placement}
The placement of the recurrent unit in the pipeline is a crucial decision. Adding it after the backbone raises the complexity of the training loop. However, we believe that feeding a lower-level data representation, only developed after the backbone, could result in better merge of scans. To verify this hypothesis, we have implemented a model similar to the one presented by FS-GRU \cite{Chen2022FS-GRU:Sharing}, where the recurrent module is fed the encoded pillars instead. Our solution outperforms in all metrics (Table \ref{tab:bef_backbone_vs}). Most noticeable is the fact that the network having the recurrent module before the backbone performs almost in the same way as the single frame baseline. 

\begin{table}[h]
\centering
\footnotesize
\begin{tabular}{c c | c c c c c}
\hline
Method & GRU & NDS & mAP & mATE & mASE & mAOE \\ \hline
PointPillars & - & 0.677 & 0.506 & 0.022 & 0.026 & 0.068 \\ \hline
TimePillars & bef. bb & 0.687 & 0.523 & 0.020 & 0.026 & 0.068 \\
TimePillars & aft. bb & \B 0.727 & \B 0.577 & \B 0.018 & \B 0.024 & \B 0.029 \\
\hline
\end{tabular}
\caption{Single-frame baseline compared to two possible recurrent configurations, differing exclusively in the location of the recurrent unit (before or after the backbone). Both methods benefit from auxiliary learning. Note all the TimePillars models used within this project employ a convGRU after the backbone, unless otherwise specified.}
\label{tab:bef_backbone_vs}
\end{table}

\subsubsection{Auxiliary task}
We also validate the adequacy of using an auxiliary task. Table \ref{tab:aux_task} confirms the extra performance achieved in all the metrics when including the auxiliary task in training.

\begin{table}[h]
\centering
\footnotesize
\begin{tabular}{c c|c c c c c}
\hline
Method & Aux. & NDS & mAP & mATE & mASE & mAOE \\ \hline
TimePillars  &  & 0.720 & 0.567 & 0.020 & 0.024 & 0.031 \\
TimePillars & \checkmark & \B 0.727 & \B 0.577 & \B 0.018 & \B 0.024 & \B 0.029 \\
\hline
\end{tabular}
\caption{Results from using or not the auxiliary task for the convolution-based transform module and using 3 sensor scans.}
\label{tab:aux_task}
\end{table}

\subsubsection{Longitudinal detection range}

As complement to the results presented, using a fixed longitudinal range of 120 m, we evaluate the performance at growing distances from the ego-vehicle. We perform distance binning, and contrast the results obtained with our approach, and with PointPillars. TimePillars outperforms the baseline, which proves to be only suitable for close objects. The benefits of considering past knowledge extend to a considerably longer detection range (see Table \ref{tab: Detect range ablation}).

\section{Conclusions}
\label{sec:conclusions}
The work shows that considering past sensor data confirms to be superior than just exploiting information at the present. Having access to previous information about the driving environment faces the sparsity nature of LiDAR point clouds, and leads to more accurate predictions. We show the adequacy of recurrent networks as a mean to achieve the latter. In contrast to point cloud aggregation methods, which consider the creation of denser data representations by heavy processing, it turns out endowing the system with memory brings a more robust solution. Our proposed method, TimePillars, materialises a way to solve the recurrent problem. 
By just augmenting the single-frame baseline with three extra convolutional layers during inference, we prove that basic network building blocks are enough to achieve significant results, and to guarantee the existing efficiency and hardware-integration specifications are met.
To the best of our knowledge, the work acts as a first baseline of results for
the 3D object detection task on the newly introduced Zenseact Open Dataset. We hope our work serves as contribution for future safer and more sustainable roads.

{\small
\bibliographystyle{ieee_fullname}

\begin{thebibliography}{10}\itemsep=-1pt

\bibitem{Alibeigi2023ZenseactDriving}
Mina Alibeigi, William Ljungbergh, Adam Tonderski, Georg Hess, Adam Lilja, Carl Lindstr{\"{o}}m, Daria Motorniuk, Junsheng Fu, Jenny Widahl, Petersson Zenseact, and Sweden Gothenburg.
\newblock {Zenseact Open Dataset: A large-scale and diverse multimodal dataset for autonomous driving}.
\newblock 5 2023.

\bibitem{Caesar2019NuScenes:Driving}
Holger Caesar, Varun Bankiti, Alex~H. Lang, Sourabh Vora, Venice~Erin Liong, Qiang Xu, Anush Krishnan, Yu Pan, Giancarlo Baldan, and Oscar Beijbom.
\newblock {nuScenes: A multimodal dataset for autonomous driving}.
\newblock {\em Proceedings of the IEEE Computer Society Conference on Computer Vision and Pattern Recognition}, pages 11618--11628, 3 2019.

\bibitem{Casas2018IntentNet:Data}
Sergio Casas, Wenjie Luo, and Raquel Urtasun.
\newblock {IntentNet: Learning to Predict Intention from Raw Sensor Data}.
\newblock 1 2018.

\bibitem{chen2022mppnet}
Xuesong Chen, Shaoshuai Shi, Benjin Zhu, Ka~Chun Cheung, Hang Xu, and Hongsheng Li.
\newblock Mppnet: Multi-frame feature intertwining with proxy points for 3d temporal object detection, 2022.

\bibitem{Chen2022FS-GRU:Sharing}
Zhikai Chen, Yafei Wang, Xulei Liu, and Xinchang Wang.
\newblock {FS-GRU: Continuous Perception and Prediction with inter Frame Feature Sharing}.
\newblock pages 517--522, 11 2022.

\bibitem{erabati2022li3detr}
Gopi~Krishna Erabati and Helder Araujo.
\newblock Li3detr: A lidar based 3d detection transformer, 2022.

\bibitem{fan2021embracing}
Lue Fan, Ziqi Pang, Tianyuan Zhang, Yu-Xiong Wang, Hang Zhao, Feng Wang, Naiyan Wang, and Zhaoxiang Zhang.
\newblock Embracing single stride 3d object detector with sparse transformer, 2021.

\bibitem{Fan2023SuperDetection}
Lue Fan, Yuxue Yang, Feng Wang, Naiyan Wang, and Zhaoxiang Zhang.
\newblock {Super Sparse 3D Object Detection}.
\newblock 1 2023.

\bibitem{DBLP:interpolation}
Artem Filatov, Andrey Rykov, and Viacheslav Murashkin.
\newblock Any motion detector: Learning class-agnostic scene dynamics from a sequence of lidar point clouds.
\newblock {\em CoRR}, abs/2004.11647, 2020.

\bibitem{Kitti}
Andreas Geiger, Philip Lenz, and Raquel Urtasun.
\newblock Are we ready for autonomous driving? the kitti vision benchmark suite.
\newblock In {\em 2012 IEEE Conference on Computer Vision and Pattern Recognition}, pages 3354--3361, 2012.

\bibitem{Huang2020AnClouds}
Rui Huang, Wanyue Zhang, Abhijit Kundu, Caroline Pantofaru, David~A Ross, Thomas Funkhouser, and Alireza Fathi.
\newblock {An LSTM Approach to Temporal 3D Object Detection in LiDAR Point Clouds}.
\newblock 7 2020.

\bibitem{Lang2018PointPillars:Clouds}
Alex~H. Lang, Sourabh Vora, Holger Caesar, Lubing Zhou, Jiong Yang, and Oscar Beijbom.
\newblock {PointPillars: Fast Encoders for Object Detection from Point Clouds}.
\newblock {\em Proceedings of the IEEE Computer Society Conference on Computer Vision and Pattern Recognition}, 2019-June:12689--12697, 12 2018.

\bibitem{Lin2017FocalDetection}
Tsung-Yi Lin, Priya Goyal, Ross Girshick, Kaiming He, and Piotr Doll{\'{a}}r.
\newblock {Focal Loss for Dense Object Detection}.
\newblock 8 2017.

\bibitem{Liu2015SSD:Detector}
Wei Liu, Dragomir Anguelov, Dumitru Erhan, Christian Szegedy, Scott Reed, Cheng-Yang Fu, and Alexander~C. Berg.
\newblock {SSD: Single Shot MultiBox Detector}.
\newblock 12 2015.

\bibitem{Loshchilov2017DecoupledRegularization}
Ilya Loshchilov and Frank Hutter.
\newblock {Decoupled Weight Decay Regularization}.
\newblock 11 2017.

\bibitem{Luo2018FastNet}
Wenjie Luo, Bin Yang, and Raquel Urtasun.
\newblock {Fast and Furious: Real Time End-to-End 3D Detection, Tracking and Motion Forecasting with a Single Convolutional Net}.
\newblock {\em Proceedings of the IEEE Computer Society Conference on Computer Vision and Pattern Recognition}, pages 3569--3577, 12 2018.

\bibitem{Mccrae3DDATA}
Scott Mccrae and Avideh Zakhor.
\newblock {3D OBJECT DETECTION FOR AUTONOMOUS DRIVING USING TEMPORAL LIDAR DATA}.
\newblock Technical report.

\bibitem{tensorRT}
NVIDIA.
\newblock Nvidia deep learning tensorrt documentation - dla layer support restrictions, 2023.
\newblock Acessed 29 August 2023.

\bibitem{Qi2016PointNet:Segmentation}
Charles~R. Qi, Hao Su, Kaichun Mo, and Leonidas~J. Guibas.
\newblock {PointNet: Deep Learning on Point Sets for 3D Classification and Segmentation}.
\newblock {\em Proceedings - 30th IEEE Conference on Computer Vision and Pattern Recognition, CVPR 2017}, 2017-January:77--85, 12 2016.

\bibitem{Qi2017PointNet++:Space}
Charles~R. Qi, Li Yi, Hao Su, and Leonidas~J. Guibas.
\newblock {PointNet++: Deep Hierarchical Feature Learning on Point Sets in a Metric Space}.
\newblock {\em Advances in Neural Information Processing Systems}, 2017-December:5100--5109, 6 2017.

\bibitem{DBLP:yolo}
Joseph Redmon, Santosh~Kumar Divvala, Ross~B. Girshick, and Ali Farhadi.
\newblock You only look once: Unified, real-time object detection.
\newblock {\em CoRR}, abs/1506.02640, 2015.

\bibitem{ren2015faster}
Shaoqing Ren, Kaiming He, Ross Girshick, and Jian Sun.
\newblock Faster {R-CNN}: Towards real-time object detection with region proposal networks.
\newblock In {\em Advances in Neural Information Processing Systems}, 2015.

\bibitem{Shi2018PointRCNN:Cloud}
Shaoshuai Shi, Xiaogang Wang, and Hongsheng Li.
\newblock {PointRCNN: 3D Object Proposal Generation and Detection from Point Cloud}.
\newblock 12 2018.

\bibitem{waymo}
Pei Sun, Henrik Kretzschmar, Xerxes Dotiwalla, Aurelien Chouard, Vijaysai Patnaik, Paul Tsui, James Guo, Yin Zhou, Yuning Chai, Benjamin Caine, Vijay Vasudevan, Wei Han, Jiquan Ngiam, Hang Zhao, Aleksei Timofeev, Scott Ettinger, Maxim Krivokon, Amy Gao, Aditya Joshi, Yu Zhang, Jonathon Shlens, Zhifeng Chen, and Dragomir Anguelov.
\newblock Scalability in perception for autonomous driving: Waymo open dataset.
\newblock {\em CoRR}, abs/1912.04838, 2019.

\bibitem{sun2022swformer}
Pei Sun, Mingxing Tan, Weiyue Wang, Chenxi Liu, Fei Xia, Zhaoqi Leng, and Dragomir Anguelov.
\newblock Swformer: Sparse window transformer for 3d object detection in point clouds, 2022.

\bibitem{tan2020efficientdet}
Mingxing Tan and Quoc~V. Le.
\newblock Efficientdet: Scalable and efficient object detection.
\newblock In {\em Proceedings of the IEEE/CVF Conference on Computer Vision and Pattern Recognition}, 2020.

\bibitem{Yan2018SECOND:Detection}
Yan Yan, Yuxing Mao, and Bo Li.
\newblock {SECOND: Sparsely Embedded Convolutional Detection}.
\newblock {\em Sensors 2018, Vol. 18, Page 3337}, 18(10):3337, 10 2018.

\bibitem{Yang2019PIXOR:Clouds}
Bin Yang, Wenjie Luo, and Raquel Urtasun.
\newblock {PIXOR: Real-time 3D Object Detection from Point Clouds}.
\newblock {\em Proceedings of the IEEE Computer Society Conference on Computer Vision and Pattern Recognition}, pages 7652--7660, 2 2019.

\bibitem{Zhou2019End-to-EndClouds}
Yin Zhou, Pei Sun, Yu Zhang, Dragomir Anguelov, Jiyang Gao, Tom Ouyang, James Guo, Jiquan Ngiam, and Vijay Vasudevan.
\newblock {End-to-End Multi-View Fusion for 3D Object Detection in LiDAR Point Clouds}.
\newblock 10 2019.

\bibitem{Zhou2017VoxelNet:Detection}
Yin Zhou and Oncel Tuzel.
\newblock {VoxelNet: End-to-End Learning for Point Cloud Based 3D Object Detection}.
\newblock {\em Proceedings of the IEEE Computer Society Conference on Computer Vision and Pattern Recognition}, pages 4490--4499, 11 2017.

\end{thebibliography}

}

\end{document}